\documentclass[10pt,twocolumn,letterpaper]{article}

\usepackage{iccv}
\usepackage{times}
\usepackage{epsfig}
\usepackage{graphicx}
\usepackage{amsmath}
\usepackage{amssymb}

\usepackage{booktabs}

\usepackage[linesnumbered,ruled]{algorithm2e}

\usepackage{newfloat}
\usepackage{listings}
\usepackage{amsfonts}
\usepackage{tabularx, wrapfig, adjustbox, multirow}

\usepackage{xcolor}

\usepackage[pagebackref=true,breaklinks=true,letterpaper=true,colorlinks,bookmarks=false]{hyperref}

\iccvfinalcopy % *** Uncomment this line for the final submission

\def\iccvPaperID{32} % *** Enter the ICCV Paper ID here

\ificcvfinal\fi

\begin{document}

%%%%%%%%% TITLE

\title{Frequency-Aware Self-Supervised Long-Tailed Learning}

% related keywords of frequency:
% density, distribution, skew, 

\author{Ci-Siang Lin$^{1,2}$ \qquad Min-Hung Chen$^{2}$ \qquad Yu-Chiang Frank Wang$^{1, 2}$ \\
$^{1}$Graduate Institute of Communication Engineering, National Taiwan University, Taiwan \\ $^{2}$NVIDIA, Taiwan\\
{\tt\small \{d08942011, ycwang\}@ntu.edu.tw, minghungc@nvidia.com}
}

% \and

\maketitle
\ificcvfinal\thispagestyle{empty}\fi

\begin{abstract}
  Data collected from the real world typically exhibit long-tailed distributions, where frequent classes contain abundant data while rare ones have only a limited number of samples. While existing supervised learning approaches have been proposed to tackle such data imbalance, the requirement of label supervision would limit their applicability to real-world scenarios in which label annotation might not be available. Without the access to class labels nor the associated class frequencies, we propose \textbf{F}requency-\textbf{A}ware \textbf{S}elf-\textbf{S}upervised \textbf{L}earning (FASSL) in this paper. Targeting at learning from unlabeled data with inherent long-tailed distributions, the goal of FASSL is to produce discriminative feature representations for downstream classification tasks. In FASSL, we first learn frequency-aware prototypes, reflecting the associated long-tailed distribution. Particularly focusing on rare-class samples, the relationships between image data and the derived prototypes are further exploited with the introduced self-supervised learning scheme. Experiments on long-tailed image datasets quantitatively and qualitatively verify the effectiveness of our learning scheme.
\end{abstract}
\vspace{-7mm}
\section{Introduction}

Deep neural networks (DNNs) have made remarkable progress in recent years and have been applied in various applications, such as image recognition~\cite{he2016deep,zheng2019looking}, semantic segmentation~\cite{badrinarayanan2017segnet,liu2019auto}, image synthesis~\cite{chen2017photographic,Meshry_2021_CVPR}, 3D analysis~\cite{Fischer_2021_CVPR,qi2017pointnet} and video understanding~\cite{Wu_2021_CVPR,zolfaghari2018eco}. When abundant labeled data are available for training (i.e., standard supervised scenario), DNNs would achieve satisfactory performance due to its powerful learning ability. However, in real-world applications, collecting labeled data is often costly and time-consuming, which thus limits the scalability and practicality of supervised DNN models. To alleviate this labeling cost, researchers start to explore varying learning strategies using \textit{unlabeled data}, resulting in the recent emergence of self-supervised learning.

Self-supervised learning (SSL)~\cite{chen2020simple,van2018representation,grill2020bootstrap,caron2020unsupervised} aims to pre-train DNN models with unlabeled data and to derive their representations, so that downstream tasks can be performed effectively (e.g., image recognition, object detection, etc.). The emergence of self-supervised learning begins from unsupervised representation learning works~\cite{gidaris2018unsupervised,noroozi2016unsupervised,zhang2016colorful}, in which pretext tasks such as rotation prediction~\cite{gidaris2018unsupervised}, jigsaw puzzle solving~\cite{noroozi2016unsupervised} or image colorization~\cite{zhang2016colorful} are designed to obtain supervision through data manipulation. As pointed out in~\cite{chen2020simple}, these pretext tasks are handcrafted and may not generalize well to downstream tasks. On the other hand, contrastive learning~\cite{chen2020simple,van2018representation} has received increasing research attention for SSL. Its core idea is to create positive pairs from different augmented views of the same image and negative pairs from different images. For instance, SimCLR~\cite{chen2020simple} combines multiple data augmentations and performs contrastive learning by attracting positive features while repelling negative ones. Despite the promising results, these works typically rely on a large number of negative pairs for training purposes. To learn without negative pairs, BYOL~\cite{grill2020bootstrap} attracts positive pairs between the student and teacher network with exponential moving average (EMA) to avoid model collapse. A recent work of SwAV~\cite{caron2020unsupervised} utilizes prototypes to perform clustering, and enforces consistent clustering assignments across positive pairs. While promising results have been presented, these SSL works are not designed to handle data with possible \textit{class imbalance} and may fail to generalize to long-tailed datasets~\cite{jiang2021self}.

Long-tailed data learning has been significant in machine learning, where frequent classes contain abundant data while rare ones have only scarce samples. Models learned from such highly imbalanced data would often be biased towards frequent classes and perform poorly on categories with relatively fewer samples. To alleviate this problem, a number of works~\cite{han2005borderline,drummond2003c4,cui2019class,park2021influence,zhu2022balanced,park2022majority} have been presented and could be divided into two categories: re-sampling and re-weighting. Re-sampling~\cite{han2005borderline,drummond2003c4,park2022majority} aims to sample instances for each class to eliminate the imbalanced issue. Over-sampling~\cite{han2005borderline,park2022majority} replicates rare-class samples, which could lead to over-fitting, while under-sampling~\cite{drummond2003c4} randomly removes frequent-class ones, which may discard valuable data. Re-weighting~\cite{cui2019class,park2021influence,zhu2022balanced}, on the other hand, assigns larger weights to rare categories during training. An intuitive way is to have weights inverse proportional to class frequencies and therefore amplify the importance of rare-class samples. Specifically, \cite{cui2019class} proposes a theoretical framework to estimate the effective number of samples and perform re-weighting. However, such methods require full label supervision to long-tailed data.

Without observing data labels during pre-training, works like~\cite{jiang2021improving,liu2021self,yang2020rethinking} investigate the performance of SSL methods on class-imbalanced or long-tailed data. A recent work of~\cite{liu2021self} also studies the improved robustness of self-supervised learning models compared to supervised ones. Assuming that rare-class samples might be ``forgotten'' by DNNs during training, SDCLR~\cite{jiang2021self} chooses to perform pruning when training the neural network, simulating the ``forgetting'' mechanism for producing DNNs which are robust to rare categories. However, since their learning mechanism treats each sample equally, the resulting model might still favor frequent classes.

To address the long-tailed data learning problem without label supervision, we propose a \textit{\textbf{F}requency-\textbf{A}ware \textbf{S}elf-\textbf{S}upervised \textbf{L}earning} (FASSL) scheme in this paper. We present a \textit{Frequency-Aware Prototype Learning} strategy in FASSL, which learns image prototypes from class-imbalanced yet unlabeled data, aiming to reflect the inherent long-tailed distribution. With such derived prototypes, we utilize a teacher-student learning scheme and present \textit{Prototypical Re-balanced Self-Supervised Learning} to train deep neural networks for producing discriminative feature representations. As noted above, this entire learning scheme does \textit{not} observe any label supervision. As confirmed later by our experiments, the frequency-aware prototypes learned by FASSL would properly describe image data with long-tailed distributions. More importantly, the CNN backbones pre-trained by FASSL can be effectively applied for downstream classification tasks, and performs favorably against state-of-the-art SSL or supervised models.

We now highlight our contributions as below:
\begin{itemize}

\item{We propose \textit{Frequency-Aware Self-Supervised Learning} (FASSL) to pre-train CNNs using unlabeled data with inherent long-tailed distributions.}

\item{In FASSL, we present a \textit{Frequency-Aware Prototype Learning} stage, which identifies frequency-aware prototypes from unlabeled data, reflecting the implicit long-tailed data distribution.}

\item{With the observed image prototypes, our \textit{Prototypical Re-balanced Self-Supervised Learning} trains CNN models from long-tailed yet unlabeled image data, benefiting downstream visual classification tasks.}

\end{itemize}

\section{Related Works}

\subsection{Self-Supervised Learning}
In real-world applications, collecting dense annotations could be costly or sometime infeasible. To alleviate such annotation costs, self-supervised learning (SSL)~\cite{chen2020simple,van2018representation,grill2020bootstrap,caron2020unsupervised} aims to learn data representations from unlabeled data. To introduce discriminative capability for the learned representation, SimCLR~\cite{chen2020simple} pulls positive samples from another augmentation view while pushes negative ones farther away from each other. In order to learn with only positive pairs while preventing models from collapse, BYOL~\cite{grill2020bootstrap} introduces student and teacher networks to attract positive pairs with exponential moving average. On the other hand, SwAV~\cite{caron2020unsupervised} assigns soft clustering codes with prototypes, and it performs swap predictions to enforce consistent clustering assignments. While promising results are presented, existing SSL methods generally assume that the training data are balanced. As verified later in our experiments, such techniques cannot generalize to long-tailed data.

\begin{figure*}[t]
    \centering
    \includegraphics[width=0.90\linewidth]{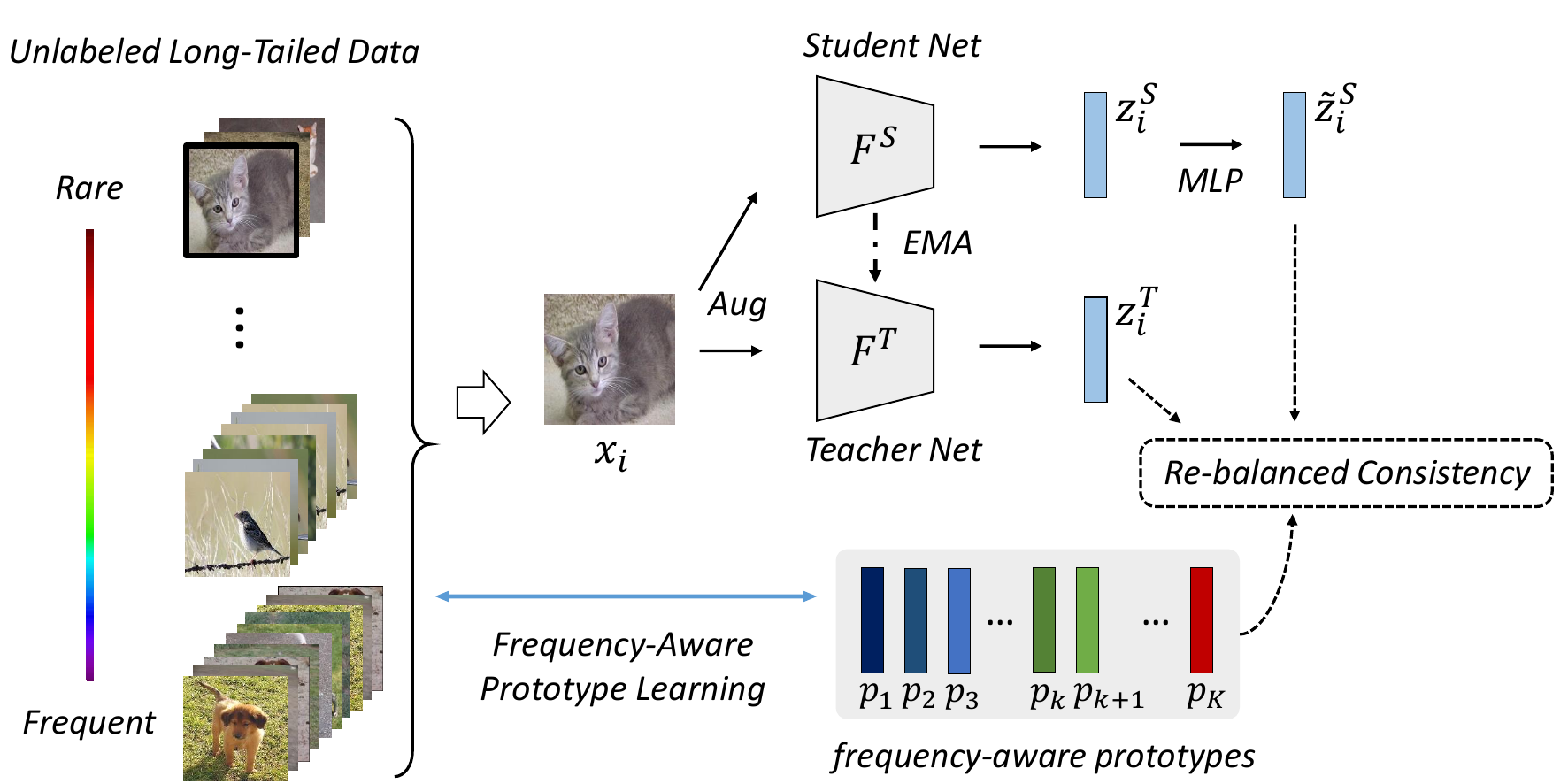}
    % \vspace{-1mm}
    \caption{Overview of the proposed \textit{\textbf{F}requency-\textbf{A}ware \textbf{S}elf-\textbf{S}upervised \textbf{L}earning} (FASSL). With image prototypes $\{p_1, p_2, \ldots, p_K\}$ derived to reflect the inherent long-tailed data distribution, each image instance is uniquely exploited into \textit{Prototypical Re-balanced Self-Supervised Learning} via a teacher-student network ($F^T$ and $F^S$) to perform representation learning.}
    \label{fig:model}
    \vspace{-2mm}
\end{figure*}
\begin{figure}[t]
    \centering
    \includegraphics[width=0.95\linewidth]{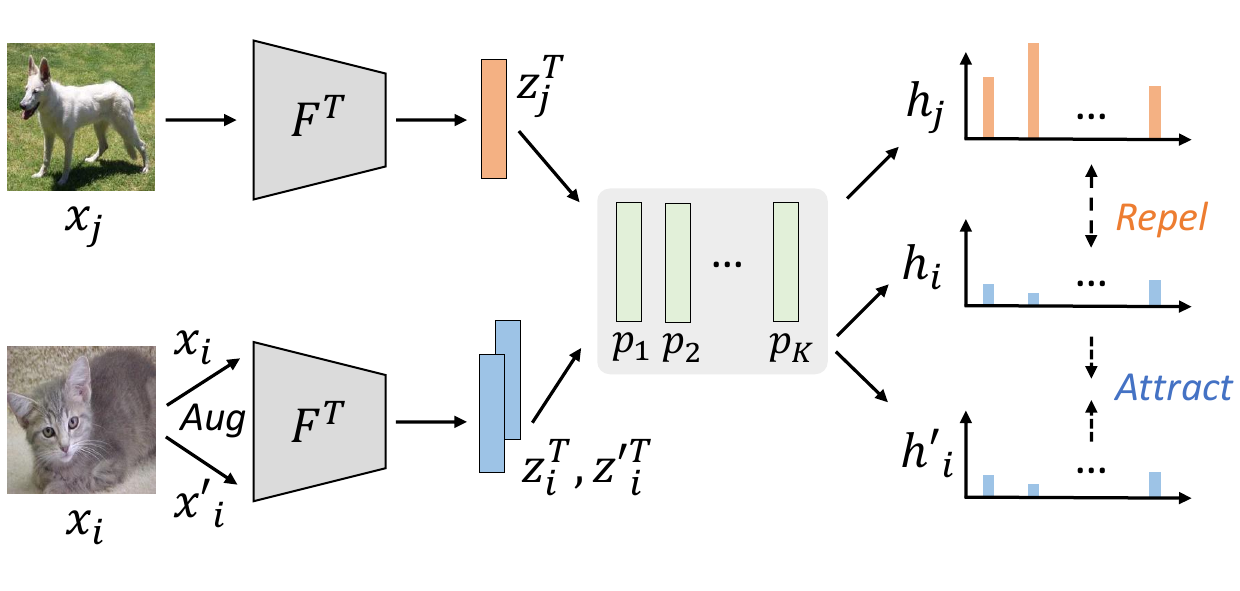}
    
    \caption{\textit{Frequency-Aware Prototype Learning}. Extended from contrastive learning, we learn image prototypes $\{p_1, p_2, \ldots, p_K\}$ from unlabeled data inputs with objectives allowing the prototypes aligning with the implicit long-tailed distribution.}
    \label{fig:model2}
    \vspace{-3.9mm}
\end{figure}

\subsection{Learning from Long-Tailed Data}

\paragraph{Supervised Long-Tailed Learning.} Real-world data typically exhibit long-tailed distributions, where head (frequent) classes contain abundant data while tail (rare) classes contain a limited amount of samples. Learning deep models which generalize well to rare classes is therefore of broad research interest. To address the imbalanced data learning problem, a number of works have been proposed~\cite{han2005borderline,drummond2003c4,cui2019class,park2021influence,li2022targeted,zhu2022balanced,park2022majority,li2021self,li2022nested,fu2022meta}. Specifically, re-sampling approaches~\cite{han2005borderline,drummond2003c4,park2022majority} design sampling strategies either to remove data from frequent classes (under-sampling) or replicate rare-class samples (over-sampling) to generate a balanced dataset. However, these methods may discard valuable data or lead to over-fitting on sampled rare-class data. On the other hand, re-weighting techniques~\cite{cui2019class,park2021influence,zhu2022balanced} weight each sample by inverse class frequencies to emphasize rare classes. Despite the promising results, these methods share a common constraint of label supervision (i.e., known label distributions). When such supervision is not available, neither strategy can be easily applied. 

\paragraph{Self-Supervised Long-Tailed Learning.} 
Without observing class labels or knowing data frequencies in advance, self-supervised long-tailed learning~\cite{bai2022effectiveness,jiang2021self,kukleva2023temperature,zhou2022contrastive} aims to pre-train deep learning models which generalize to rare categories. Specifically, COLT~\cite{bai2022effectiveness} alleviates class imbalance by adding additional training data sampled from an auxiliary dataset. When such external data are not available, SDCLR~\cite{jiang2021self} emphasizes the learning on rare samples by performing model pruning and enforcing a consistency loss between the pruned model and the original one. On the other hand, BCL~\cite{zhou2022contrastive} designs data augmentation with different intensities and applies stronger augmentation to data with higher training losses. However, take BCL~\cite{zhou2022contrastive} 
 as an example, it requires defining a large set of specific augmentation types by heuristic to achieve satisfactory performance.
\section{Proposed Method}
\subsection{Notations and Overview}

We first provide the problem definition and the notations used in this paper. Assume there is a set of $N$ images $X = \{x_i\}_{i=1}^{N}$ with the associated imbalanced label set $Y = \{y_i\}_{i=1}^{N}$, where $x_i \in \mathbb{R}^{H \times W \times 3}$ and $y_i \in \mathbb{R}$ represent the $i^{th}$ image and its corresponding class label, respectively. Following the standard SSL setting, we do \textit{not} observe the label set $Y$ during pre-training, while $X$ is expected to exhibit a long-tailed distribution. It is worth repeating that, without label supervision, existing approaches for imbalanced data such as re-sampling~\cite{han2005borderline,drummond2003c4,park2022majority} and re-weighting~\cite{cui2019class,park2021influence,zhu2022balanced} cannot be directly applied.

In this paper, we propose \textit{\textbf{F}requency-\textbf{A}ware \textbf{S}elf-\textbf{S}upervised \textbf{L}earning} (FASSL) for pre-training deep learning models using unlabeled long-tailed data. As illustrated in Figure~\ref{fig:model}, we exploit the inherent imbalanced data distribution \textit{without} any label supervision during the training stage, which allows us to derive discriminative representation for downstream classification tasks. We propose to learn frequency-aware prototypes $\{p_1, p_2, \ldots, p_K\}$ from unlabeled data, which reflect the long-tailed data distribution. With such prototypes obtained, we design self-supervised objectives for learning the CNN backbone model. Note that we have $F^T$ and $F^S$ indicate the teacher and student networks in our framework (parameterized by $\theta^T$ and $\theta^S$, respectively). We now detail our proposed method in the following subsections.

\subsection{Frequency-Aware Self-Supervised Learning}

\subsubsection{Frequency-Aware Prototype Learning}
\label{sec:pcl}

To start our proposed learning strategy, we first utilize unlabeled long-tailed data to derive image representatives, aiming to reflect the inherent \textit{imbalanced data distributions}. We view this as the learning stage of \textit{Frequency-Aware Prototype Learning}, which advances contrastive learning technique to produce the desirable image prototypes $\{p_1, p_2, \ldots, p_K\}$. We now detail this learning stage. 

In order to derive frequency-aware prototypes describing long-tailed data distribution in an unsupervised setting, we first follow SimCLR~\cite{chen2020simple} and perform data augmentation on sampled input images $x_i$ to create positive pairs and extract the image features $z^T_i$ and ${z'}^T_i$ with the network $F^T$. 
We note that, rather than directly imposing the contrastive loss on image features $z^T_i$ as SimCLR did, we choose to perform contrastive learning on the \textit{similarity score distribution} level $h_i$, which is derived from the inner products between $x_i$ and $\{p_1, p_2, \ldots, p_K\}$. Since such inner product operation suggests the similarity between each image and the prototypes, we expect that such derived prototypes would learn as visual exemplars for long-tailed data.

More specifically, we form the prototype matrix $P$ by taking the prototypes in each row, and the resulting matrix size is $K \times D$, indicating $K$ $D$-dimensional prototypes. We perform matrix-vector product from the matrix $P$ and the feature $z^T_i$ to produce the similarity scores $h_i$ (which  could be implemented with a linear layer):
\begin{equation}
\begin{aligned}
h_i = P\cdot{z}^T_i, \quad \text{where} \quad {z}^T_i = F^T(x_i)
\end{aligned}.
\label{eq:proto}
\end{equation}

And, a standard contrastive loss $L_{contra}$ is calculated as:
\begin{equation}
\begin{aligned}
\label{eq:loss_contra}
    L_{contra} = \mathbb{E}_{x_i\sim X}\left[-\log \frac{\exp(sim(h_i, h'_i)/\beta)}{\sum_{j}\exp(sim( h_i, h_j)/\beta)}\right]
\end{aligned},
\vspace{1mm}
\end{equation}
where $sim(\cdot, \cdot)$ denotes the cosine similarity and $\beta$ is a hyperparameter of temperature. From the above design, the image prototypes are encouraged to align $h_i$ with its positive sample $h'_i$ (derived from ${z'}^T_i$) while repelling negative ones $h_j$. As rare-class data are less frequently sampled and learned, the prototypes are more likely to be updated and to describe frequent categories. As a result, the resulting prototypes would be expected to exhibit the inherent long-tailed distribution.

We note that, both the \textit{network} $F^T$ and \textit{prototypes} $\{p_1, p_2, \ldots, p_K\}$ will be updated by $L_{contra}$ via back propagation. Once complete, the resulting prototypes would implicitly describe the long-tailed data distribution (i.e., a large portion of $\{p_1, p_2, \ldots, p_K\}$ would correspond to frequent classes, while only few are associated with the rare classes). Later in the experiments, we will verify and visualize the derivation of image prototypes at this stage.

\subsubsection{Prototypical Re-balanced SSL}
\label{sec:reb}
As pointed out by~\cite{jiang2021self}, existing SSL approaches like~\cite{chen2020simple,van2018representation,grill2020bootstrap,caron2020unsupervised} are vulnerable to class imbalance, which might fail to generalize to long-tailed data problems. To address this particular challenge, our FASSL utilizes the frequency-aware prototypes derived above as a guidance for learning discriminative representations from unlabeled long-tailed data, as discussed below.

Our proposed strategy is to perform self-supervised learning from unlabeled data with inherent data distribution utilized, while jointly achieving the goal of training the network $F^S$ for producing discriminative feature representations. Given the aforementioned prototypes implicitly reflecting long-tailed data distribution, we would expect different degrees of similarity when relating each sampled image to these prototypes, depending on its corresponding class frequency. To be more precise, given an input image $x_i$, we first extract its feature $z^T_i$ from the teacher network $F^T$ (initialized from Section~\ref{sec:pcl}). If $x_i$ is from a rare class, it would be expected to be dissimilar to a majority of prototypes and therefore be viewed as more important during the SSL process. As a result, as a reweighting technique, we calculate the weight $\phi(z^T_i, P)$ for $x_i$, which is inversely proportional to the similarity sum to all prototypes. 

With the above reweighting and regularization strategy, we adopt the asymmetric teacher-student framework to perform \textit{Prototypical Re-balanced Self-Supervised Learning}, as depicted in Figure~\ref{fig:model}. To avoid possible model collapse when attract positive pairs~\cite{grill2020bootstrap}, we further deploy a MLP for the student network $F^S$. It can be seen that we encourage the semantic similarity between $\tilde{z}^S_i$ and $z^T_i$ derived from $F^S$ and $F^T$ with the re-balanced self-supervised consistency loss $L_{reb}$:
\vspace{-1mm}
\begin{equation}
\begin{aligned}
\label{eq:loss_reb}
L_{reb} =\mathbb{E}_{x_i\sim X}\left[\phi(z^T_i, P)\cdot L_{consis}(\tilde{z}^S_i, z^T_i)\right],\\[12pt]
\text{where} \quad \phi(z^T_i, P) = \frac{1}{exp(\sum_{k}{\langle z^T_i, p_k \rangle)}} \\[8pt]
\text{and} \quad L_{consis} = \left\lVert \frac{\tilde{z}^S_i}{\left\lVert\tilde{z}^S_i\right\rVert_2} - \frac{z^T_i}{\left\lVert z^T_i\right\rVert_2}\right\rVert_2^2.
\end{aligned}
\end{equation}
Thus, the student network $F^S$ would be updated by the above re-balanced loss $L_{reb}$, while the teacher network $F^T$ is updated by exponential moving average (EMA) from $F^S$:
\begin{equation}
\label{eq:ema}
\begin{aligned}
\theta_{S} \leftarrow \theta_{S} - \gamma \frac{\partial L_{reb}}{\partial \theta_{S}}\quad \text{and}\\
\theta_{T} \leftarrow \tau\cdot\theta_{T} + (1-\tau)\cdot\theta_{S},
\end{aligned}
\vspace{1mm}
\end{equation}
where $\tau$ controls the decay rate of the teacher network $F^T$, and $\theta_{S}$ and $\theta_{T}$ are parameters of $F^S$ and $F^T$. It can be seen that, with the prototypes derived in Section~\ref{sec:pcl} and the introduced re-balanced loss $L_{reb}$, we are able to identify rare-class samples and focus on the associated representation learning. It is also worth repeating that, the above learning scheme is implemented with no label supervision.

Finally, we note that we do not alternate between the above two stages during training. This is because, alternative optimization between these two stages tends to hinder the prototypes from describing the associated long-tailed distribution. With $F^T$ initialized by SDCLR~\cite{jiang2021self}, we simply jointly train $F^T$ and produce the prototypes $\{p_1, p_2, \ldots, p_K\}$ in \textit{Frequency-Aware Prototype Learning}. As for the stage of \textit{Prototypical Re-balanced Self-Supervised Learning}, we perform the proposed re-weighted SSL to learn the network $F^S$ for data with long-tailed distribution. The pseudo code for our FASSL is summarized in Algorithm~\ref{algo}.
\begin{algorithm}[t]

\textbf{Input}: Images $X = \{x_i\}_{i=1}^{N}$ with long-tailed class distribution

\BlankLine

\textbf{\textit{Frequency-Aware Prototype Learning}}

$\{p_1, p_2, \ldots, p_K\}$ $\leftarrow$ randomly initialize

$\theta_{T}$ $\leftarrow$ initialize from ~\cite{jiang2021self}

\For{num. of iterations}{

$x_i$, $x'_i$, $x_j$ $\leftarrow$ random sample from $X$

$h_i$, $h'_i$, $h_j$ $\leftarrow$ derived by~(\ref{eq:proto}) with $F^T$ and $P$

$L_{contra}$ $\leftarrow$ calculated by~(\ref{eq:loss_contra})

$\theta_{T}$, $p_1, p_2, \ldots, p_K$ $\leftarrow$ update by $L_{contra}$
}

\BlankLine

\textbf{\textit{Prototypical Re-balanced Self-Supervised Learning}}

$\theta_{S}$ $\leftarrow$ initialize from $\theta_{T}$

\For{num. of iterations}{

$x_i$, $x'_i$ $\leftarrow$ random sample from $X$

$\tilde{z}^S_i$, $z^T_i$ $\leftarrow$ produced by $F^T$ and $F^S$

$L_{reb}$ $\leftarrow$ calculated by~(\ref{eq:loss_reb})

$\theta_{S}$, $\theta_{T}$ $\leftarrow$ update by $L_{reb}$ and EMA~(\ref{eq:ema})
}

\BlankLine

\textbf{Output}: the student network $F^S$
\caption{Training of FASSL}
\label{algo}
\normalsize
% \vspace{-5mm}
\end{algorithm}
% \vspace{-5mm}

\begin{table*}[t]
\centering
% \vspace{5mm}

% \vspace{3mm}
\caption{Evaluation on Tiny-ImageNet, ImageNet-100-LT, CIFAR100-LT and CIFAR10-LT with the standard setting (i.e., use of all labeled data for fine-tuning). $\uparrow$~denotes the higher the better, and $\downarrow$~denotes the lower the better. \textbf{Bold} denotes the best averaged results except for the supervised method. \text{\textdagger}: Note that BYOL~\cite{grill2020bootstrap} can be viewed as an ablation study of ours (\ie, teacher-student learning with an uniform-weighting loss).}
\label{tab:standard}
    \vspace{2mm}
    \begin{adjustbox}{width=0.83\textwidth}
    \begin{tabular}{llc|ccccc}
    \toprule
    Dataset   & Method & All $\uparrow$ &  \textit{Rare} $\uparrow$ & \textit{Medium} $\uparrow$ & \textit{Frequent} $\uparrow$ & \textit{Std} $\downarrow$  \\ \midrule

    \multirow{6}{*}{Tiny-ImageNet}
    &   SimCLR~\cite{chen2020simple} & 32.11 & 32.12 & 31.73 & 32.47 & 0.30  \\
    &   SwAV~\cite{caron2020unsupervised}  & 33.21 & 32.15 & 30.61 & 36.88 & 2.67 \\
    &   BYOL\text{\textdagger}~\cite{grill2020bootstrap} & 37.58 & 34.91 & 35.45 & 42.38 & 3.40 \\
    &   SDCLR~\cite{jiang2021self} & 45.23 & 42.15 & 43.39 & 50.15 & 3.51  \\
    &   \textbf{FASSL (Ours)} & \textbf{45.85} & 43.15 & 43.88 & 50.53 & 3.32 \\
    \cmidrule{2-7}
    &   Supervised  & 43.57 & 40.00 & 40.45 & 50.26 & 4.74 \\

    \midrule

    \multirow{3}{*}{ImageNet-100-LT}
    &   SimCLR~\cite{chen2020simple} & 65.46 & 59.69 & 63.71 & 69.54 & 4.04  \\
    % &   SwAV~\cite{caron2020unsupervised}  & 33.21 & 32.15 & 30.61 & 36.88 & 2.67 \\
    % &   BYOL\text{\textdagger}~\cite{grill2020bootstrap} & 37.58 & 34.91 & 35.45 & 42.38 & 3.40 \\
    &   SDCLR~\cite{jiang2021self} & 66.48 & 60.92 & 65.04 & 70.10 & 3.75   \\
    % &   \textbf{FASSL (Ours)} & \textbf{45.85} & 43.15 & 43.88 & 50.53 & 3.32 \\
    &   \textbf{FASSL (Ours)} & \textbf{68.92} &  66.00 & 68.06 & 72.71 & 2.81  \\
    % \cmidrule{2-7}
    % &   Supervised  & \textcolor{red}{?} & \textcolor{red}{?} & \textcolor{red}{?} & \textcolor{red}{?} & \textcolor{red}{?} \\

    \midrule
    \multirow{8}{*}{CIFAR100-LT}
    &   LDAM-DRW+SSP~\cite{yang2020rethinking} & 43.43 & - & - & - & -\\
    &   SwAV~\cite{caron2020unsupervised}  & 47.00 & 44.00 & 47.03 & 49.97 & 2.44  \\
    &   BYOL\text{\textdagger}~\cite{grill2020bootstrap} & 48.86 & 45.55 & 47.42 & 53.62 & 3.45  \\
    &   SimCLR~\cite{chen2020simple}  & 49.76 & 47.58 & 50.36 & 51.35 & 1.60 \\
    &   BCL-I~\cite{zhou2022contrastive} & 52.22 & 48.27 & 53.03 & 55.35 & 2.95 \\
    &   SDCLR~\cite{jiang2021self} & 54.94 & 51.00 & 55.03 & 58.79 & 3.18 \\
    
    % &   BCL-I~\cite{zhou2022contrastive} & 55.23 & 51.64 & 56.15 & 57.91 & 2.64 \\
    &    \textbf{FASSL (Ours)} & \textbf{55.27} & 53.55 & 54.52 & 57.74 & 1.79  \\
    \cmidrule{2-7}
    &   Supervised  & 54.06 & 51.39 & 54.18 & 56.62 & 2.13 \\

    \midrule
    
    \multirow{7}{*}{CIFAR10-LT}
    &   SimCLR~\cite{chen2020simple} & 75.37 & 69.33 & 73.33 & 83.45 & 5.94  \\
    &   BYOL\text{\textdagger}~\cite{grill2020bootstrap} & 75.66 & 75.43 & 69.83 & 81.70 & 4.85 \\
    &   SwAV~\cite{caron2020unsupervised} & 76.60 & 73.30 & 71.10 & 85.40 & 6.29  \\
    &   LDAM-DRW+SSP~\cite{yang2020rethinking} & 77.83 & - & - & - & -\\
    &   SDCLR~\cite{jiang2021self} & 80.49 & 75.10 & 78.07 & 88.30 & 5.66  \\
    &    \textbf{FASSL (Ours)} & \textbf{80.69} & 78.80 & 76.73 & 86.55 & 4.23  \\
    \cmidrule{2-7}
    &   Supervised & 80.76 & 75.43 & 80.93 & 85.93 & 4.95  \\

    \bottomrule
    \centering
    % \vspace{-2mm}
    
    \end{tabular}
    
    \end{adjustbox}
    
\end{table*}

\begin{table*}[ht!]
\centering

\caption{Evaluation on CIFAR100-LT/CIFAR10-LT with the few-shot setting (i.e., use of $1\%$ of labeled data for fine-tuning). \text{\textdagger}: Note that BYOL~\cite{grill2020bootstrap} could be viewed as an ablation study of ours (\ie, teacher-student learning with an uniform-weighting loss).}
\label{tab:few}
    \vspace{3mm}
    \begin{adjustbox}{width=0.75\textwidth}
    \begin{tabular}{llc|ccccc}
    \toprule
    Dataset   & Method & All $\uparrow$  & \textit{Rare} $\uparrow$ & \textit{Medium} $\uparrow$ &  \textit{Frequent} $\uparrow$ & \textit{Std} $\downarrow$ \\ \midrule

    \multirow{6}{*}{CIFAR100-LT}
    &   SwAV~\cite{caron2020unsupervised}  & 20.14 & 13.91  & 20.06 & 26.44 & 5.12 \\
    &   BYOL\text{\textdagger}~\cite{grill2020bootstrap} & 20.62 & 14.03 & 19.61 & 28.24 & 5.84 \\
    &   SimCLR~\cite{chen2020simple} & 22.51 & 15.45  & 22.48 & 29.59 & 5.77 \\
    &   SDCLR~\cite{jiang2021self} & 25.39 & 19.91 & 25.52 & 30.74 & 4.42 \\
    &   \textbf{FASSL (Ours)}  & \textbf{27.11} & 21.18 & 27.88 & 32.26 & 4.56 \\
    
    \cmidrule{2-7}
    &   Supervised & 28.97 & 13.88 & 32.03 & 41.00 & 11.28 \\
    
    \midrule
    
    \multirow{6}{*}{CIFAR10-LT}
    &   BYOL\text{\textdagger}~\cite{grill2020bootstrap} & 62.33 & 50.40 & 60.53 & 76.05 & 10.55 \\
    &   SimCLR~\cite{chen2020simple}  & 62.98 & 56.67 & 62.97 & 69.30 & 5.16\\
    &   SwAV~\cite{caron2020unsupervised}  & 65.16 & 52.90 & 61.97 & 80.60 & 11.53 \\
    &   SDCLR~\cite{jiang2021self}  & 67.23 & 56.63 & 61.87 & 83.20 & 11.49 \\
    &   \textbf{FASSL (Ours)} & \textbf{68.54} & 62.43  & 61.93 & 81.25 & 8.99 \\
    \cmidrule{2-7}
    &   Supervised & 69.82 & 65.07 & 66.47 & 77.93 & 5.76 \\

    \bottomrule
    \centering
    \vspace{-5mm}
    \end{tabular}
    \end{adjustbox}
    
\end{table*}

\section{Experiments}

\subsection{Experimental Settings}

\paragraph{Datasets.} Following~\cite{jiang2021self} and~\cite{cui2019class}, we consider the benchmarks of Long-Tailed CIFAR-10/CIFAR-100 and Long-Tail ImageNet-100 for experiments. The original CIFAR-10/CIFAR-100~\cite{krizhevsky2009learning} datasets contain $10$/$100$ classes in a total of $60,000$ $32\times32$ images. To create the long-tailed setting,~\cite{cui2019class} samples imbalanced subsets from the originals to create Long-Tailed CIFAR-10/CIFAR-100. The imbalance factor is defined by the number of samples in the most frequent class divided by the least one. Moreover, we follow SDCLR~\cite{jiang2021self} and set the imbalance factor as $100$, which makes the imbalanced problem challenging. For both datasets, we randomly choose one of the five splits for the experiments. As for Long-Tail ImageNet-100, the sample number per class ranges
from $1280$ to $5$. In addition, we also evaluate the proposed method on Tiny-ImageNet~\cite{le2015tiny}. Tiny-ImageNet contains $200$ classes and each class contains $500$ training images. We sample an imbalanced pre-training subset from the training split for the long-tailed setting and take the validation split as the testing data.

\paragraph{Settings and Evaluation.} We consider the standard linear evaluation in self-supervised learning to measure the quality of learned representations. That is, we freeze the parameters of the backbone model and fine-tune a linear classifier on top of it. Since our student network $F^S$ contains a ResNet-18~\cite{he2016deep} as the CNN backbone model and an additional MLP as the projection head, we  follow SimCLR~\cite{chen2020simple} and remove the projection head during fine-tuning. In addition to standard pre-train/fine-tune setting, we follow SDCLR~\cite{jiang2021self} and consider the \textit{few-shot} setting, where only $1\%$ data in the standard setting are used to fine-tune the classifier. Depending on the sample frequencies, we divide all classes into three groups, \textit{Frequent}, \textit{Medium} and \textit{Rare}. \textit{Frequent} stands for the top thrid of most frequent classes, while \textit{Rare} stands for the lowest third. We report the accuracy of each group and also the average and standard deviation of three groups.

\subsection{Implementation Details}

We perform data augmentation of random cropping, horizontal flipping, color jittering and random gray scaling for pre-training, with ResNet-18/50~\cite{he2016deep} as the backbone. We set the temperature $\beta$ as $0.2$ and the number of prototypes $K$ as $128$ by default. We clip outliers and normalize the weights in each mini-batch for stability. For fine-tuning, we use the standard cross-entropy loss and train $30$/$100$ epochs for the standard and few-shot setting, respectively.

\begin{table*}[t]
    \centering

    \caption{Hyper-parameter analysis on the number of prototypes $K$ (\textbf{left}) and the decay rate $\tau$ of EMA (\textbf{right}).}
    \label{tab:hyper}
    \vspace{3mm}
    \begin{adjustbox}{width=0.98\textwidth}
    
    \begin{tabular}{l|ccccc|ccccc}
    \toprule
    Dataset & $K$ & All $\uparrow$ & \textit{Rare} $\uparrow$ & \textit{Medium} $\uparrow$ & \textit{Frequent} $\uparrow$ & $\tau$ & All $\uparrow$  & \textit{Rare} $\uparrow$ & \textit{Medium} $\uparrow$ & \textit{Frequent} $\uparrow$\\ 
    \midrule

    \multirow{5}{*}{CIFAR100-LT} 
    & 64 & 54.94 & 52.67 & 54.30 & 57.85              & 0.0 & 33.35 & 30.15 & 34.88 & 35.03 \\
    & 96 & 54.99 & 52.15 & 54.24 & 58.59              & 0.9 & 53.63 & 50.91 & 52.97 & 57.00 \\
    & 128 & \textbf{55.27} & 53.55 & 54.52 & 57.74    & 0.99 & \textbf{55.27} & 53.55 & 54.52 & 57.74 \\
    & 256 & 54.99 & 52.15 & 54.88 & 57.94             & 0.999 & 55.02 & 51.70 & 55.00 & 58.35 \\
    & 512 & 55.22 & 52.88 & 54.61 & 58.18             & 1.0 & 53.00 & 50.73 & 53.24 & 55.03 \\
    
    \midrule
    
    \multirow{5}{*}{CIFAR10-LT} 
    & 64 & 80.69 & 78.43 & 75.33 & 88.30              & 0.0 & 49.74 & 44.27 & 47.83 & 57.13\\
    & 96 & \textbf{80.81} & 77.07 & 77.20 & 88.15     & 0.9 & 79.74 & 75.40 & 76.57 & 87.25   \\
    & 128 & 80.69 & 78.80 & 76.73 & 86.55           & 0.99 & \textbf{80.69} & 78.80  & 76.73 & 86.55\\
    & 256 & 80.79 & 78.97 & 75.53  & 87.88             & 0.999 & 80.16 & 77.57 & 74.30 & 88.63 \\
    & 512 & 80.69 & 77.13 & 76.70 & 88.05             & 1.0 & 79.63 & 75.33 & 75.83 & 87.73 \\

    \midrule
    
    \multirow{5}{*}{Tiny-ImageNet} 
    & 64 & 45.80 & 43.06 & 43.76 & 50.59             & 0.0 & 30.70 & 25.61 & 27.61 & 38.88 \\
    & 96 & 45.50 & 43.03 & 43.42 & 50.06             & 0.9 & \textbf{45.86} & 42.76 & 44.12 & 50.71 \\
    & 128 & 45.85 & 43.15  & 43.88 & 50.53           & 0.99 & 45.85 & 43.15 & 43.88 & 50.53 \\
    & 256 & \textbf{46.11} & 43.18 & 44.09 & 51.06   & 0.999 & 45.75 & 42.91 & 43.94 & 50.41 \\
    & 512 & 45.72 & 42.97 & 43.70 & 50.50            & 1.0 & 45.83 & 42.48 & 44.48 & 50.53 \\
    
    \bottomrule
    \end{tabular}
    \end{adjustbox}
    \centering
    % \vspace{-1mm}
    
\end{table*}

\subsection{Quantitative Comparisons}

In Table~\ref{tab:standard}, we perform linear evaluation on Long-Tailed CIFAR-10/CIFAR-100, Long-Tail ImageNet-100 and Tiny-ImageNet. 
 We note that, while no label supervision during pre-training, neither SimCLR~\cite{chen2020simple} nor SwAV~\cite{caron2020unsupervised} consider the long-tailed or imbalanced setting. 
 From the results shown in Table~\ref{tab:standard}, we see that our approach achieved the best performance compared to existing self-supervised methods. Specifically, we achieved the accuracy of $45.85\%$ in average and $43.15\%$ on rare classes on Tiny-ImageNet. Since our approach explicitly weighted rare samples higher to tackle class imbalance, we performed favorably against SDCLR~\cite{jiang2021self} by $1\%$ on rare classes. It is worth noting that, since BYOL~\cite{grill2020bootstrap} adopts teacher-student learning with a uniform-weighting loss, it could be viewed as an ablation study of our re-balanced loss $L_{reb}$. We see that BYOL only reported $34.91\%$ on rare classes, which is over $8\%$ lower compared to our method, verifying the effectiveness of our loss $L_{reb}$ on unlabeled long-tailed data. We also note that, BCL-I~\cite{zhou2022contrastive} requires defining a large set of specific augmentation types by heuristic. When only common augmentation types in SSL are applied (as we do), BCL-I reported the rare-class accuracy of $48.27\%$ on CIFAR100-LT, which is over $5\%$ lower compared to our FASSL. In addition, our method is even \textit{higher} than supervised learning on CIFAR100-LT, which further verifies the effectiveness of our \textit{Frequency-Aware Self-Supervised Learning} framework.

\begin{figure}[t]
    \centering
    \includegraphics[width=1.0\linewidth]{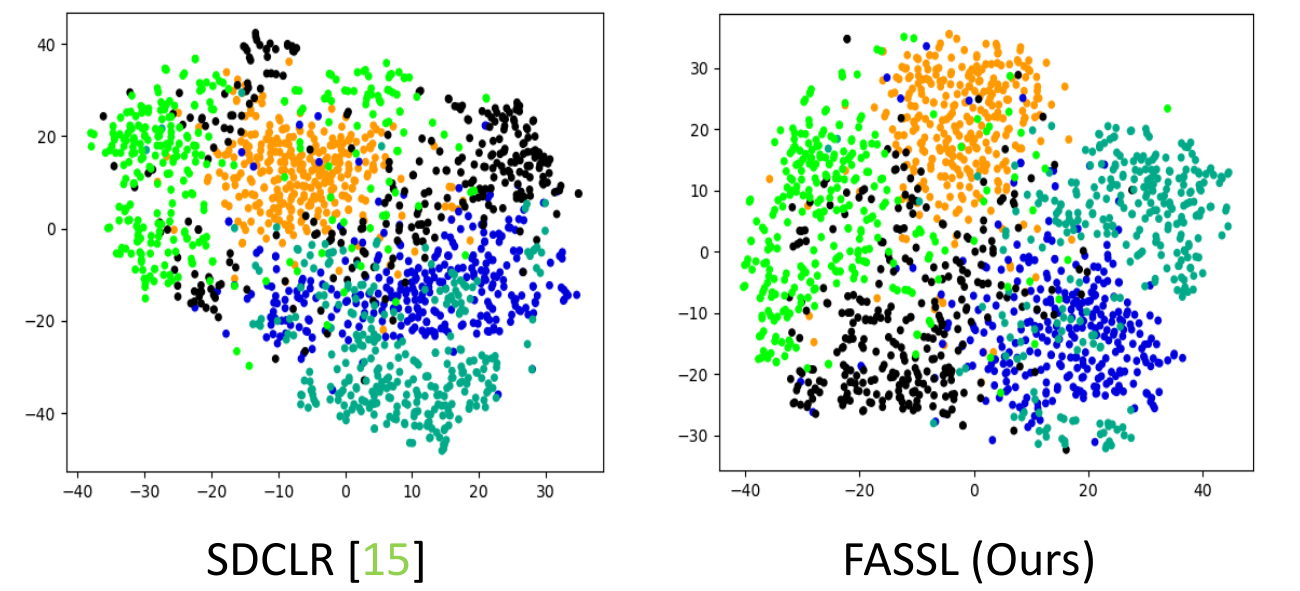}
    \vspace{-4mm}
    \caption{T-SNE of latent representations on CIFAR10-LT. Different colors indicate different categories. Compared with SDCLR~\cite{jiang2021self}, our FASSL results in improved discriminative representations.}
    \label{fig:tsne}
   % \vspace{-5mm}
\end{figure}

We further consider the challenging few-shot setting, in which only $1\%$ labeled data are used to train the linear classifier. This would assess the model ability in deriving robust representations and avoiding over-fitting on few labeled data. In Table~\ref{tab:few}, we observe that our model again performed favorably against existing methods under the few-shot setting. In particular, we achieved $62.43\%$ on rare-classes on CIFAR10-LT, which is over $5\%$ higher compared to SDCLR~\cite{jiang2021self}. To further verify the effectiveness of our approach, we provide qualitative comparisons as shown in Figure~\ref{fig:tsne}. We select five categories from CIFAR10-LT and apply t-SNE visualization for the learned features of the student network $F^S$. We see that the representations derived by our scheme are better separated according to the associated categories.

\subsection{Analysis of Hyper-parameters}

\paragraph{Impact of the number of prototypes $K$.} In Table~\ref{tab:hyper} (left), we conduct sensitivity analysis with respect to the number of prototypes $K$. When varying $K$ from $64$ to $512$ on CIFAR100-LT, we see that our model produced consistent results within a $0.5\%$ drop compared to the best average accuracy. It is worth noting that, different from other SSL works~\cite{yue2021prototypical} which utilize prototypes to perform clustering, we do \textit{not} require knowing the number of classes in advance to set the number of prototypes $K$. Thus, we simply set $K$ to 128 throughout our experiments.

\vspace{-3mm}
\begin{figure*}[t]
    \centering
    \includegraphics[width=1.0\linewidth]{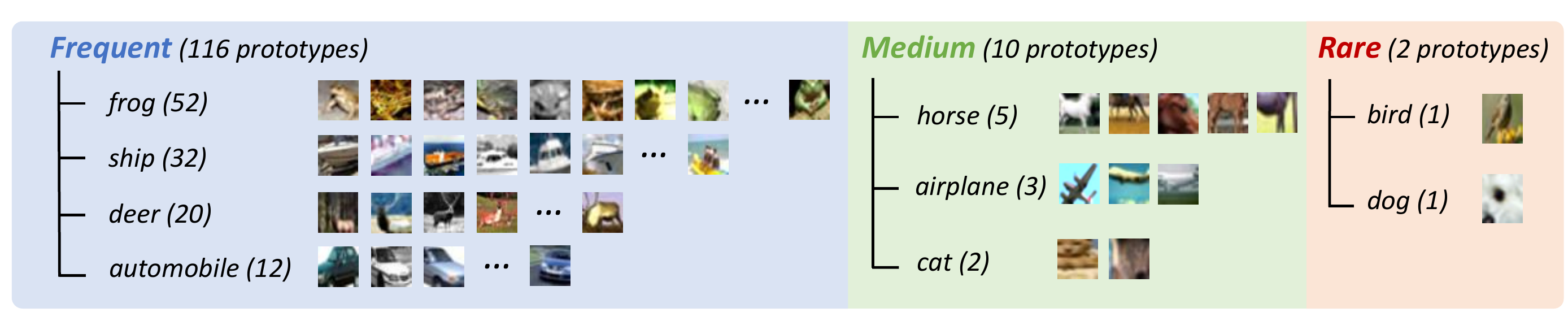}
    % \vspace{0.05mm}
    \caption{Visualization of 128 frequency-aware prototypes and the associate classes on CIFAR10-LT. The prototypes are visualized by retrieving their closest images in the latent space. Note that the number in each parenthesis denotes the number of prototypes in each group/category.}
    \vspace{-2mm}
    \label{fig:proto}
\end{figure*}

\paragraph{Impact of the decay rate $\tau$ of EMA.} In our proposed FASSL, we adopt exponential moving average (EMA) to update the teacher network $F^T$ with the decay rate $\tau$. We now analyze the impact of $\tau$ in Table~\ref{tab:hyper} (right). When setting $\tau=0$, no moving average is adopted, and the teacher network $F^T$ shares the same parameters with the student network $F^S$ at all times. As noted in~\cite{grill2020bootstrap}, attracting positive pairs between identical feature encoders could lead to model collapse and therefore yielded the poor performance of $33.35\%$ on CIFAR100-LT. When the decay rate $\tau=1$, the teacher network $F^T$ is frozen and never updated, resulting in suboptimal results. In practice, we set $\tau=0.99$ to progressively distill the knowledge from the student network $F^S$ while avoiding collapse.

\begin{table}[t]
    \centering

    \caption{Different numbers of prototypes $K$ and the corresponding \textbf{class distributions} on Tiny-ImageNet, CIFAR100-LT and CIFAR10-LT with the imbalanced factor $\rho$ set as 100.}
    \label{tab:freq}
    \vspace{3mm}
    \begin{adjustbox}{width=0.43\textwidth}
    \begin{tabular}{lcccc}
    \toprule
    Dataset & $K$ & \textit{Frequent} & \textit{Medium} & \textit{Rare} \\ 
    \midrule
    
    \multirow{4}{*}{Tiny-ImageNet} 
    & 64 & 78.13\% & 21.87\% & 0.00\%  \\
    & 96 & 86.46\% & 13.54\% & 0.00\%  \\
    & 128 & 84.38\% & 14.06\% & 1.56\%  \\
    & 256 & 84.77\% & 13.67\% & 1.56\% \\
    & 512 & 89.26\% & 9.77\% & 0.97\%  \\
    
    \midrule
    
    \multirow{4}{*}{CIFAR100-LT} 
    & 64 & 89.06\% & 7.81\%  & 3.13\% \\
    & 96 & 82.29\% & 15.63\%  & 2.08\% \\
    & 128 & 88.28\% & 10.16\%  & 1.56\%  \\
    & 256 & 85.94\% & 12.89\% & 1.17\% \\
    & 512 & 91.21\% & 6.84\% & 1.95\% \\
    % & 512 & 50.10 & 47.78 & 43.36 & 3.09 & 47.11\\
    
     \midrule
     
    \multirow{4}{*}{CIFAR10-LT} 
    & 64 & 90.63\% & 9.38\%  & 0.00\% \\
    & 96 & 90.63\% & 8.33\%  & 1.04\% \\
    & 128 & 90.63\% & 7.81\%  & 1.56\%  \\
    & 256 & 88.28\% & 9.38\% & 2.34\% \\
    & 512 & 85.35\% & 12.50\% & 2.15\% \\
    
    \bottomrule
    \end{tabular}
    \end{adjustbox}
    \vspace{-3mm}
\end{table}

\vspace{1mm}
\subsection{Analysis of Long-Tailed Prototypes}
In \textit{Frequency-Aware Prototype Learning}, we learn prototypes to reflect the long-tailed distribution from unlabeled data. With the imbalance factor $\rho=100$, the amount of frequent samples is about ten times of that of the medium ones (same for medium vs. rare classes). Therefore, most of the prototypes are expected to represent frequent classes. To confirm this, we take $128$ prototypes to retrieve closest images in the latent space and visualize them in Figure~\ref{fig:proto}. We see that a total of $116$ prototypes were from frequent-class images (e.g., frog, ship, etc.), and only $2$ were related to rare classes of bird and dog on CIFAR10-LT. 

In addition to visualization, we also present the class distributions of the retrieved images in Table~\ref{tab:freq}. With $K = 64$ on CIFAR10-LT, we observe that the number of prototypes may not be sufficient to cover rare classes. With $K = 128$ and above, the distributions were aligned with the long-tailed data. Similar results are observed on CIFAR100-LT and Tiny-ImageNet. In Table~\ref{tab:imbalance}, we further vary the imbalanced factor $\rho$ from $1$ to $500$ to control the degree of class imbalance and report the corresponding class distributions of prototypes on Tiny-ImageNet. We see that when $\rho=1$ indicating no class imbalance presents, the class distribution was nearly uniform as desired. With larger $\rho$, the prototypes number ratio of frequency classes over rare class was increased. This validates that our prototypes effectively describe the data distribution, and thus are applicable for our \textit{Prototypical Re-balanced Self-Supervised Learning}.

\begin{table}[t]
    \centering
    
    % \vspace{3mm}
    \caption{\textbf{Class distributions} of prototypes when varying imbalanced factor $\rho$ $=1-500$.}
    \vspace{3mm}
    \label{tab:imbalance}
    \begin{adjustbox}{width=0.43\textwidth}
    \begin{tabular}{lcccc}
    
    \toprule
    
    Dataset & $\rho$ & \textit{Frequent} & \textit{Medium} & \textit{Rare} \\ 
    
    \midrule
    
    \multirow{5}{*}{Tiny-ImageNet} 
    & 1 & 34.38\% & 33.59\% & 32.03\% \\
    & 20 & 72.66\% & 21.09\% & 6.25\% \\
    & 100 & 84.38\% & 14.06\% & 1.56\%  \\
    & 200 & 87.50\% & 10.94\% & 1.56\% \\
    & 500 & 91.41\% & 7.81\% & 0.78\% \\

    \bottomrule
    
    \end{tabular}
    % \vspace{-mm}
    \end{adjustbox}
    \vspace{-3mm}
    
\end{table}

\section{Conclusion}
Learning representations from long-tailed data has been among active research topics for machine learning communities, and it becomes particularly challenging when no label supervision is available during training (or pre-training) stage. In this paper, we proposed \textit{\textbf{F}requency-\textbf{A}ware \textbf{S}elf-\textbf{S}upervised \textbf{L}earning} (FASSL), which exploits and identifies the inherent imbalanced data distribution to derive frequency-aware prototypes. While self-supervised learning can be easily deployed on the unlabeled input data, our FASSL further utilizes the aforementioned image prototypes for guiding the learning process, which aligns with the data distributions while producing desirable image representations. Once the network backbone is pre-trained by FASSL, downstream  classification tasks can be tackled with satisfactory performances. Experiments on long-tailed benchmarks confirmed the effectiveness of our FASSL against state-of-the-art methods, while ablation studies and analyses were conducted to verify and visualize our derived prototypes and representations.

\vspace{-2mm}

\paragraph{Acknowledgment}
This work is supported in part by the National Science and Technology Council under grant NSTC-111-2634-F-002-020 and National Taiwan University under grant NTU-112L900901. We also thank to National Center for High-performance Computing (NCHC) for providing computational and storage resources.

{\small
\bibliographystyle{ieee_fullname}
\bibliography{egbib}
}

\clearpage

\begin{table}[t]
    \centering
    \caption{Evaluation on Places-LT with the standard setting (i.e., use of all labeled
data for fine-tuning).}
    \vspace{2mm}
    \label{tab:placeslt}
    
    \begin{adjustbox}{width=0.46\textwidth}
    \begin{tabular}{lc|cccc}
    \toprule
    Method & All & \textit{Rare} & \textit{Medium} & \textit{Frequent} & \textit{Std}\\ 
    \midrule
    SDCLR~[15] & 21.50  & 7.18  & 18.58  & 38.74  & 13.05 \\
    FASSL (Ours) & \textbf{22.89}  & \textbf{7.98}  & \textbf{20.35}  & \textbf{40.34}  & 13.34 \\
    \bottomrule
    \end{tabular}
    \end{adjustbox}
    % \vspace{-2mm}
\end{table}
\begin{table}[t]
    \centering
    \caption{Performance of our proposed FASSL with/without alternate training on CIFAR100-LT.}
    \vspace{2mm}
    \label{tab:alternate}
    
    \begin{adjustbox}{width=0.47\textwidth}
    \begin{tabular}{lc|ccc}
    \toprule
    Method & All   & \textit{Rare} & \textit{Medium} & \textit{Frequent}\\ 
    \midrule
    FASSL (w/ alternate training) & 55.06  & 52.58 & \textbf{54.94} & 57.68\\
    FASSL (w/o alternate training) & \textbf{55.27}  & \textbf{53.55} & 54.52 & \textbf{57.74} \\
    \bottomrule
    \end{tabular}
    \end{adjustbox}
    % \vspace{-2mm}
\end{table}

\section{Additional Experiments}

\subsection{Experiments on Places-LT}

Places-LT is a long-tailed dataset sampled from
Places~[A]. It contains $365$ categories with a total of $62,500$ images. The amount of data in each class ranges
from $4,980$ to $5$. In Table~\ref{tab:placeslt}, we show that SDCLR~[15] achieved the accuracy of $21.50\%$ while our FASSL reported $\textbf{22.89\%}$ on Places-LT and is therefore preferable.

\subsection{Ablation Studies}

\paragraph{Alternate Training.}
To address the long-tailed data learning problem without label supervision, we propose a \textit{Frequency-Aware Self-Supervised Learning} (FASSL) scheme, which is composed of two learning stages: \textit{Frequency-Aware Prototype Learning} and \textit{Prototypical Re-balanced Self-Supervised Learning}. In Table~\ref{tab:alternate}, we demonstrate the results of our FASSL with or without alternating between the above two learning stages. We see that alternate training would result in degraded performance. This is because alternate optimization tends to hinder the prototypes from describing long-tailed data distributions (and also increases the training time). Therefore, we choose not to alternate between the two stages.

\begin{table}[t]
    \centering
    \caption{Performance of our proposed FASSL when using different model architectures on CIFAR100-LT.}
    \vspace{2mm}
    \label{tab:model}
    
    \begin{adjustbox}{width=0.46\textwidth}
    \begin{tabular}{lcc|ccc}
    \toprule
    Method & Model & All & \textit{Rare} & \textit{Medium} & \textit{Frequent}\\ 
    \midrule
    FASSL & ResNet-$34$ & 54.19 & 51.45 & \textbf{54.88} & 56.24\\
    FASSL & ResNet-$18$ & \textbf{55.27} & \textbf{53.55}  & 54.52 & \textbf{57.74}\\
    \bottomrule
    \end{tabular}
    \end{adjustbox}
    % \vspace{-2mm}
\end{table}

\paragraph{Different Model Architectures.} In Table~\ref{tab:model}, we provide experimental results and show that deeper CNN models (e.g., ResNet-$34$) are not preferable on CIAFR100-LT due to possible overfitting problems. Thus, we choose to use ResNet-$18$ on CIAFR100-LT as existing works~[15] did.

\paragraph{Model Initialization.} In Table~\ref{tab:initialization}, we observe that if we train our model from scratch without any initialization, FASSL only achieves $71.87\%$ on rare categories. This is because our FASSL performs \textit{data-distribution-level} contrastive learning instead of \textit{image-level} one to identify the imbalanced data distribution, and therefore image-level patterns/features may not be well captured. To address this issue, we choose to initialize our CNN model from image-level SSL methods, and we see that the rare-class accuracy would improve to $72.70\%$ and $78.80\%$ when initialized from SimCLR~[5] and SDCLR~[15], respectively. This demonstrates that when using any image-level SSL methods for initialization, our FASSL consistently improves the performance on long-tailed data.

\begin{table}[!t]
    \centering
    % \vspace{-1mm}
    \caption{Performance of our proposed FASSL with different initialization on CIFAR10-LT.}
    \vspace{2mm}
    \label{tab:initialization}
    
    \begin{adjustbox}{width=0.39\textwidth}
    \begin{tabular}{lcccc}
    \toprule
    Method & All & \textit{Rare} \\
    \midrule
    FASSL (w/o initialization)  & 76.11 & 71.87 \\
    \midrule
    SimCLR~[5] & 75.37 & 69.33 \\
    FASSL (init. from SimCLR)  & 76.42 & 72.70 \\
    
    \midrule
    SDCLR~[15] &  80.49 & 75.10 \\
    FASSL (init. from SDCLR)   & \textbf{80.69} & \textbf{78.80} \\
    \bottomrule
    \end{tabular}
    \end{adjustbox}
    % \vspace{-2mm}
\end{table}

\begin{table}[t]
    \centering
    \caption{Comparison with semi-supervsied learning works when using 30\% labeled data on CIFAR100-LT.}
    \vspace{2mm}
    \label{tab:semi}
    
    \begin{adjustbox}{width=0.3\textwidth}
    \begin{tabular}{l|c}
    \toprule
    Method & All \\
    \midrule
    FixMatch w/ CReST+~~[B]  & 42.0  \\
    FASSL (Ours) & \textbf{52.1} \\
    \bottomrule
    \end{tabular}
    \end{adjustbox}
    % \vspace{-2mm}
\end{table}

\subsection{Comparison with Semi-Supervised Learning Works}

Since labeled data is also required in linear evaluation phase (i.e., finetuning a linear classifier), we also compare our method with semi-supervised works~[B], as shown in Table~\ref{tab:semi}. By using the same amount ($30\%$) of labeled data, our FASSL achieved the averaged accuracy of $\textbf{52.1\%}$ while [B] only reported $42.0\%$ on CIFAR100-LT. Thus, the use of our scheme to properly weigh and regularize long-tailed data for SSL would be desirable.

\vspace{2mm}
\vspace{11mm}

\noindent [A] Zhou, B., Lapedriza, A., Khosla, A., Oliva, A., \& Torralba, A. (2017). Places: A 10 million image database for scene recognition. IEEE transactions on pattern analysis and machine intelligence, 40(6), 1452-1464.

\noindent [B] Wei, C., Sohn, K., Mellina, C., Yuille, A., \& Yang, F. (2021). Crest: A class-rebalancing self-training framework for imbalanced semi-supervised learning. In Proceedings of the IEEE/CVF conference on computer vision and pattern recognition (pp. 10857-10866).
\end{document}